\documentclass[twocolumn,10pt]{IEEEtran}
\usepackage{algorithm,algorithmic,amsbsy,amsmath,amssymb,epsfig,bbm,mathrsfs,multirow,amsthm,subcaption,caption,graphicx,epstopdf,multicol,lipsum,float}
\usepackage{adjustbox}

\newenvironment{myalign}{\par\nobreak\scriptsize\noindent\align}{\endalign}

\begin{document}

\title{Optimal Stochastic Delivery Planning in Full-Truckload and Less-Than-Truckload Delivery}

\author{\IEEEauthorblockN
{Suttinee Sawadsitang\IEEEauthorrefmark{1}, 
 Rakpong Kaewpuang\IEEEauthorrefmark{1},
 Siwei Jiang\IEEEauthorrefmark{2}, 
 Dusit Niyato\IEEEauthorrefmark{1}},
 Ping Wang\IEEEauthorrefmark{1}\\
\IEEEauthorblockA{
\IEEEauthorrefmark{1} School of Computer Science and Engineering, Nanyang Technological University\\
\IEEEauthorrefmark{2}Singapore Institute of Manufacturing Technology (SIMTech) A*STAR } \vspace{-5mm}	}

\maketitle

\begin{abstract}
With an increasing demand from emerging logistics businesses, Vehicle Routing Problem with Private fleet and common Carrier (VRPPC) has been introduced to manage package delivery services from a supplier to customers. However, almost all of existing studies focus on the deterministic problem that assumes all parameters are known perfectly at the time when the planning and routing decisions are made. In reality, some parameters are random and unknown. Therefore, in this paper, we consider VRPPC with hard time windows and random demand, called Optimal Delivery Planning (ODP). The proposed ODP aims to minimize the total package delivery cost while meeting the customer time window constraints. We use stochastic integer programming to formulate the optimization problem incorporating the customer demand uncertainty. Moreover, we evaluate the performance of the ODP using test data from benchmark dataset and from actual Singapore road map. 
\end{abstract}



\section{Introduction}


Recently, new businesses such as franchise and online shopping have been introduced and become popular rapidly. However, these businesses introduce sophisticated logistics requirements for product/package delivery. Given an increase in delivery destinations and demand fluctuation, suppliers need to strategize their delivery plan to minimize cost while meeting customer demand. Meanwhile, two delivery modes exist. The first delivery mode is for the supplier to rent a whole truck and use the truck for package delivery. This is referred to as {\em full-truckload (FTL)} delivery mode.\footnote{``FTL",``Fleet", ``Private vehicle", and ``Truck" are interchangeable.} The second mode is for the supplier to outsource the delivery to a third-party carrier in an on-demand basis. This is referred to as {\em less-than-truckload (LTL)} delivery mode.\footnote{``LTL" and ``Carrier" are interchangeable.} Suppliers can use either FTL, LTL, or the combination to achieve their business goal. 



The FTL, i.e., renting a truck, is likely to be cheaper than LTL if the truck capacity is fully or almost fully utilized. However, truck renting requires an advance reservation at which the supplier may not know the exact customer demands. The actual demand can be more or less than the reserved truck capacity, which result in under- and over-reservation problems, respectively. Furthermore, a customer may specify a time window of delivery. The time window is a period of time that a package must be delivered. Time window constraints can be hard and soft. In hard time window constraint, the delivery must be done within the window. In soft time window constraint, the delivery can be outside the window, but this incurs penalty cost to the delivery. With the time window constraints, the supplier can use on-demand LTL delivery services from a third party carrier.

To address the aforementioned challenges, we consider the vehicle planning and routing problem to design optimal delivery routes from a depot to a number of geographically scattered customers~\cite{ref_greensurvey}. In particular, we aim to obtain the routes for full-truckload (FTL) vehicles from a single depot to the customers to which their packages have to be delivered. The delivery must be in the time window of each customer. Alternatively, the LTL service can be used by the supplier in which the carrier is responsible to find the delivery route by itself. In this paper, we propose the Optimal Delivery Planning (ODP) to minimize the total delivery cost for suppliers. As such, we formulate an optimization problem based on stochastic integer programming. Uncertainty in the customer demands is taken into account together with customer time window constraints. The stochastic integer programming model allows to analyze and optimize FTL vehicle planning and routing and LTL services on multiple time stages given the available information of the uncertainty. Moreover, we perform extensive performance evaluation of the ODP based on both standard benchmark and actual test data of Singapore road map. Compared with the baseline methods, the proposed ODP can achieve significantly lower total delivery cost.

\section{Related Work}

Vehicle Routing Problem (VRP) is one of the major research topics in supply chain management. The survey by Canhong Lin {\em et al}~\cite{ref_greensurvey} (2014) shows the significantly increasing interests of VRP from researchers in academia and practitioners in industry due to new emerging business opportunities such as online shopping. However, most of the related work considered only private truckloads or full-truckload (FTL). Meanwhile, Vehicle Routing Problem with Private fleet and common Carrier (VRPPC) has got only little attention. VRPPC has some benefits over classical VRP as it allows more flexibility for the suppliers to use less-than-truckload (LTL) for delivery. However, VRPPC is more complex than VRP as external carriers are involved.

VRPPC has been introduced in~\cite{ref_1983}. The VRPPC was formulated using mathematical programming to obtain the solutions regarding (i) which size of the FTL truck will be applied, (ii) which customer should be served by the FTL truck or an LTL carrier, and (iii) what the routing path of the FTL truck will be. To obtain the solutions, the greedy algorithm is adopted. The VRPPC~\cite{ref_1983} has been extended since then, e.g., \cite{ref_TL-LTL},~\cite{ref_SRI},~\cite{ref_2opt},~\cite{ref_genertic}. 

The authors in~\cite{ref_TL-LTL} formulated VRPPC by using integer programming~\cite{ref_TL-LTL}. They added initial cost of FTL trucks and cost of LTL carrier into objective function. The decision variable of LTL carriers was added into the optimization model, and the LTL decision variable was solved together with FTL and routing path. They proposed a heuristic algorithm, called TL-LTL algorithm to obtain the solution. The authors in~\cite{ref_SRI} proposed the improved heuristic algorithm of~\cite{ref_TL-LTL}, called ``Selection, Routing and Improvement (SRI)''. The authors in~\cite{ref_SRI} used $\lambda$-interchange procedure in the result improvement step while the authors in~\cite{ref_TL-LTL} used three procedures, i.e., intra-route two-exchanges, inter-route one-exchange and two-exchanges. The results from SRI algorithm are always closer to optimal values than TL-LTL results. The authors in~\cite{ref_2opt} introduced the Iterated Density Estimation Evolutionary Algorithm (IDEA). It was shown that IDEA performs better than SRI. The comparison between genetic, SRI, and TL-LTL algorithms were presented in~\cite{ref_genertic}. The evaluation indicated that under various settings the genetic and SRI algorithms can reach optimal results more frequently than the TL-LTL algorithm. Furthermore, the genetic algorithm obtains a solution faster than the TL-LTL algorithm.

Although there are different optimization formulations and algorithms for VRPPC, we consider different aspects of the problem. In particular, as the customer demand is not known exactly when the supplier makes a decision in reality, existing formulations and algorithms which rely heavily on this information are not applicable. Therefore, we introduce the two-stage stochastic integer programming model to address the uncertainty issue of the customer demand. Moreover, we consider the time window constraint and truck traveling limit constraints as well as the improved subtour elimination. The comparison between the ODP and existing algorithms i.e., TL-LTL, SRI, IDEA, is not presented as they did not consider randomness in the formulation. 


\section{System Model and Assumptions}
\begin{figure}[h]
\begin{center}
$\begin{array}{c} 
 \includegraphics[clip,trim=0cm 20cm 2cm 0cm,width=0.8\textwidth]{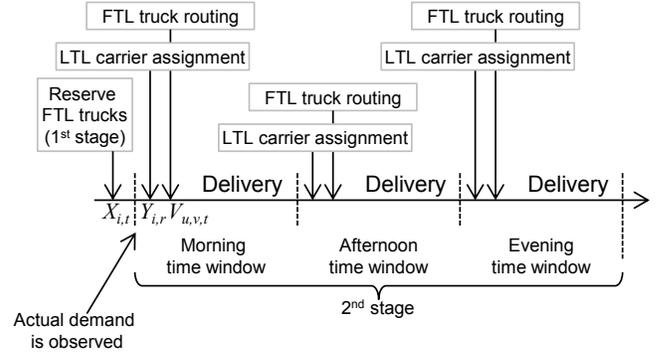} 
\end{array}$
\caption{Timing diagram of decision making in Optimal Delivery Planning (ODP).}
\label{fig:stages}
\end{center}
\end{figure}

In this section, we describe the Optimal Delivery Planning (ODP) system model. With the last mile delivery, we consider the decisions in the ODP to be made in two stages (Figure~\ref{fig:stages}). The first stage is when the supplier decides how many and which FTL trucks will be used to deliver packages to which customers. The decision will be made based on the available FTL truck information and the probability distribution of customer demand. Here, the exact customer demand has not been known yet. In the second stage, the supplier makes two decisions after the actual demand is observed. The first decision is to determine the customers to be served by a particular FTL truck or by LTL service from a carrier. The second decision is to find the best delivery route of the FTL trucks to the assigned customers. 

The supplier has a set of customers to deliver packages. Let $\mathcal{C} = \{C_1, C_2, \dots, C_{n'}\}$ denote the set of customers, where $n'$ denotes the total number of customers. Let $\omega = (D_1, D_2, \dots, D_{n'})$ be a scenario of all customers. $D_i$ represents a binary parameter of demand from customer $i$ in which $D_i=1$ means that customer $C_i$ has demand, i.e., there is a package to be delivered, and $D_i=0$ otherwise. Note that two or more packages for the same customer can be grouped as one package. $\omega$ can be regarded as a scenario in which the set of scenarios is denoted by $\mathcal{Q}$, i.e., $\omega \in \mathcal{Q}$. For example, if the supplier has three customers, the demand scenario is denoted by $\omega = (D_1, D_2, D_3)$ in which $(1,1,0)$ indicates that customers 1 and 2 have the demand while customer 3 does not. Let $A = ( A_1, A_2, \dots , A_{n'} ) $ be a list of package weights, the unit of which is kilogram. We assume that if customer $i$ has demand, i.e., $D_i=1$, then the package of the demand has the weight of $A_i$.

Let $\mathcal{T} = \{T_1, T_2, \dots, T_{t'}\}$ denote a set of FTL trucks of the supplier, where $t'$ is the total number of trucks. Let $\mathcal{R} = \{R_1, R_2, \dots, R_{r'}\}$ denote a set of LTL carriers, where $r'$ is the total number of carriers. The supplier has to determine the route of FTL trucks starting from the depot to visit all assigned customers, and returning to the depot after the delivery is completed. Let $\mathcal{U} = \mathcal{C} \cup \{\text{depot}\}$ denote a set of locations of customers and the depot, which is the source and the sink of the routing. The distance from location $u$ to location $v$ is denoted by $K_{u,v}$.

\subsection{Customer Time Windows}

Time window is a time period that a customer requires its package to be delivered. In this paper, we consider three time windows, i.e., morning, afternoon, and evening. Let $\mathcal{I}_m$, $\mathcal{I}_a$, and $\mathcal{I}_e$ be the sets of customers who will receive their packages in the morning, afternoon, and evening, respectively ($\mathcal{I}_m, \mathcal{I}_a, \mathcal{I}_e \subset \mathcal{C} $ and $\mathcal{I}_m\cup \mathcal{I}_a\cup \mathcal{I}_e = \mathcal{C}$). The sets $\mathcal{I}_m$, $\mathcal{I}_a$, and $\mathcal{I}_e$ must not include the same customers ($\mathcal{I}_m \cap \mathcal{I}_a = \emptyset$, $\mathcal{I}_m \cap \mathcal{I}_e = \emptyset$, $\mathcal{I}_a \cap \mathcal{I}_e = \emptyset$). For each time window, a route of each FTL truck must not exceed its distance limit. The traveling distance limits are denoted as $L^{(m)}$, $L^{(a)}$, and $L^{(e)}$ for morning, afternoon, and evening, respectively. In addition, we assume that the package dispatching time at each customer is short and negligible. 

\subsection{Pricing}

Three different payments are considered. 
\begin{itemize}
	\item $\bar{C}_t$ denotes the initial cost for FTL truck $t$. The initial cost can include driver stipend, truck rental fee, and maintenance expense. 
	\item $\widehat{C}_{i,r}$ denotes the LTL carrier service charge of delivering a package to customer $i$ by carrier $r$. 
	\item $\ddot{C}_{u,v}$ denotes the routing cost of the FTL truck from location $u$ to location $v$, where $u,v \in \mathcal{U}$. The routing costs are calculated based on the distance between~$u$~and~$v$.
\end{itemize}
	




\section{Problem Formulation}

We formulate the stochastic integer programming model for the ODP. There are five decision variables in the model.
\begin{itemize}
	\item $X_{i,t}$ is an FTL truck allocation in which $X_{i,t}=1$ if FTL truck $t$ is allocated to customer $i$, and $X_{i,t}=0$ otherwise.
	\item $Y_{i,r}$ is an LTL carrier allocation in which $Y_{i,r}=1$ if LTL carrier $r$ is used for customer $i$, and $Y_{i,r}=0$ otherwise.
	\item $W_t$ is a variable indicating the use of FTL truck for the delivery in which $W_t=1$ if FTL truck $t$ will be used, and $W_t=0$ otherwise.
	\item $V_{u,v,t}$ is a routing variable in which $V_{u,v,t} = 1$ if FTL truck $t$ will travel from location $u$ to location $v$. Again, $u,v=0$ represents the depot.
	\item $S_{i,t}$ is an auxiliary variable for eliminating a subtour in the routing solution. 
\end{itemize}

The objective function given in (\ref{e_obj1}) and (\ref{eq_obj2}) is to minimize total payment which includes (i) initial cost of FTL truck, (ii) LTL carrier service charge, and (iii) the cost of FTL vehicle routing. The expressions in (\ref{e_obj1}) and (\ref{eq_obj2}) represent the first stage and second stage objectives, respectively. In the second stage, $P(\omega)$ is the probability of scenario $\omega$. The term $ \sum_{i \in \mathcal{C}}\sum_{t \in \mathcal{T}}X_{i,t} $ is used to minimize the allocation of FTL trucks to customers. 

\vspace{1em}
\noindent Minimize: 
\begin{myalign}
\label{e_obj1}
\begin{split}
& \sum_{i \in \mathcal{C}}\sum_{t \in \mathcal{T}}X_{i,t} 
+ \sum_{t\in \mathcal{T}}\bar{C}_tW_t + E[\mathscr{Q}( X_{i,t}(\omega), W_t (\omega)  )]	,	
\end{split}
\end{myalign}
{\normalsize where}
\begin{myalign}
\begin{split}
 \mathscr{Q}( X_{i,t}(\omega), W_t (\omega) ) = 
& \sum_{i\in \mathcal{C}}\sum_{r\in \mathcal{R}}\sum_{\omega \in \mathcal{Q}}
P(\omega)\widehat{C}_{i,r}Y_{i,r}(\omega) + \\
& \sum_{u\in \mathcal{U}}\sum_{v \in \mathcal{U}}\sum_{t\in \mathcal{T}}\sum_{\omega \in \mathcal{Q}}	P(\omega)\ddot{C}_{u,v}V_{u,v,t} (\omega)	
\label{eq_obj2}
\end{split} 
\end{myalign}
subject to: (3) - (19)
\vspace{1em}

\begin{myalign}
&\sum_{t\in \mathcal{T}}X_{i,t} + \sum_{r \in \mathcal{R}}Y_{i,r}(\omega) \geq D_i(\omega), & \forall i \in \mathcal{C}, \forall \omega \in \mathcal{Q}	\label{eq_assign} \\
&\sum_{i\in \mathcal{C}}A_iX_{i,t} \leq F_t, 	\label{eq_capacity}
& \forall t \in \mathcal{T} \\
&\sum_{i\in \mathcal{C}}X_{i,t} \leq 1000W_t, 
& \forall t \in \mathcal{T}\label{eq_ftl}\\
&V_{u,u,t}(\omega) = 0
&\forall	u\in \mathcal{U}, \forall t\in \mathcal{T}, \forall \omega \in \mathcal{Q} \label{e_startend}\\
&\sum_{u \in \mathcal{U}} V_{u,0,t}(\omega) \leq 1, 
&\forall t\in \mathcal{T}, \forall \omega \in \mathcal{Q}\label{e_from0}\\
&\sum_{u \in \mathcal{U}} V_{0,u,t}(\omega)\leq 1, 
& \forall t\in \mathcal{T}, \forall \omega \in \mathcal{Q}\label{e_to0}\\
&\sum_{u \in \mathcal{U}} V_{u,i,t}(\omega)= X_{i,t}D_i(\omega) & \forall i \in \mathcal{C}, \forall t\in \mathcal{T}, \forall \omega \in \mathcal{Q} \label{e_from}\\
&\sum_{u \in \mathcal{U}} V_{i,u,t}(\omega) = X_{i,t}D_i(\omega) & \forall i \in \mathcal{C} ,\forall t\in \mathcal{T}, \forall \omega \in \mathcal{Q}\label{e_to}\\
&\sum_{u \in \mathcal{ U}}\sum_{i\in \mathcal{I}^{(m)}}V_{u,i,t}(\omega)K_{u,i} \leq L^{(m)}& \forall t \in \mathcal{T}, \forall \omega \in \mathcal{Q} \label{e_limit_m}\\
&\sum_{u \in \mathcal{U}}\sum_{i\in \mathcal{I}^{(a)}}V_{u,i,t}(\omega)K_{u,i} \leq L^{(a)}& \forall t \in \mathcal{T}, \forall \omega \in \mathcal{Q} \label{e_limit_a}\\
&\sum_{u \in \mathcal{U}}\sum_{i\in \mathcal{I}^{(e)}}V_{u,i,t}(\omega)K_{u,i} \leq L^{(e)}& \forall t \in \mathcal{T}, \forall \omega \in \mathcal{Q} \label{e_limit_e}
\end{myalign}
\vspace{-1em}
\begin{myalign}
&S_{i,t}(\omega) - S_{j,t}(\omega) + |\mathcal{C}|V_{i,j,t}(\omega) \leq |\mathcal{C}|-1,\nonumber \\
& \qquad\qquad\qquad\qquad\qquad\qquad\qquad \forall i,\forall j \in \mathcal{C}, \forall t \in \mathcal{T}, \forall \omega \in \mathcal{Q} \label{e_subtour}\\
&S_{i,t}(\omega)D_i(\omega) \leq S_{j,t}(\omega)+|\mathcal{I}^{(m)}|(1-D_i(\omega)),\nonumber \\
&\qquad\qquad\qquad \qquad\qquad\forall i \in \mathcal{I}^{(m)}, \forall j\in \mathcal{I}^{(a)}, \forall t \in \mathcal{T}, \forall \omega \in \mathcal{Q} \label{e_order1}\\
&S_{i,t}(\omega)D_i(\omega) \leq S_{j,t}(\omega)+|\mathcal{I}_{(a)}|(1-D_i(\omega)),\nonumber\\
& \qquad\qquad\qquad\qquad\qquad\forall i \in \mathcal{I}^{(a)}, \forall j\in \mathcal{I}^{(e)}, \forall t \in \mathcal{T}, \forall \omega \in \mathcal{Q} \label{e_order2}
\end{myalign}
\vspace{-2em}

\begin{myalign}
&X_{i,t},W_{t} \in \{0,1\},& \forall i \in \mathcal{C},\forall t \in \mathcal{T} \label{e_bound1}\\
&Y_{i,r}(\omega),V_{u,v,t}(\omega), \in \{0,1\},& r \in \mathcal{R}, \forall u, \forall v \in \mathcal{U}, \forall t \in \mathcal{T}, \forall \omega \in \mathcal{Q} \label{e_bound2}\\
&S_{i,t}(\omega) \in \{0,1,2,3,\dots,n'\},& \forall i \in \mathcal{C},\forall t \in \mathcal{T}, \forall \omega \in \mathcal{Q} \label{e_bound3}
\end{myalign}
The constraint in (\ref{eq_assign}) ensures that if customers have demand, their packages will be assigned to one FTL truck or one LTL carrier. For each FTL truck, the constraint in (\ref{eq_capacity}) ensures that the weight of packages must not exceed the capacity of FTL truck $t$ denoted by $F_t$. The initial cost of each FTL truck $t$ must be paid when any demand is assigned to FTL truck $t$ as indicated in the constraint in (\ref{eq_ftl}). We assume that the total number of customers is less than 1000. 

The constraints in (\ref{e_startend})-(\ref{e_to}) are to find an optimal routing path. The constraint in (\ref{e_startend}) eliminates all the paths that start and end at the same point. The constraints in (\ref{e_from0})-(\ref{e_to0}) are for the depot constraints. In particular, they guarantee that FTL trucks will have one path departing from and returning to the depot. Similar to the depot constraints, the constraints in (\ref{e_from})-(\ref{e_to}) are for the customers to be delivered by the FTL truck. In this case, the routing of customer $C_i$ will not be selected, i.e., $V_{u,i,t}, V_{i,u,t} = 0$, if customer $C_i$ does not have demand, or the customer has demand but the demand is not assigned to FTL truck $t$.

To control the delivery distance limit of each FTL truck, the constraints in (\ref{e_limit_m}), (\ref{e_limit_a}), and (\ref{e_limit_e}) are used for morning, afternoon, and evening time windows, respectively. For example, each FTL truck cannot travel more than 100 kilometres in the morning, e.g., due to speed limit regulations. 

The subtour elimination constraint is given in (\ref{e_subtour}). $S_{i,t}(\omega)$ indicating a visiting order of routing if the path from and to customer $i$ by truck $t$ is selected. For each FTL truck, this constraint does not allow two or more disjointed tours to cover all assigned customers. One example of FTL truck $t$ serving five customers is $depot \rightarrow C_5 \rightarrow C_3 \rightarrow C_1 \rightarrow C_2	\rightarrow C_4 \rightarrow depot$, where $S_{1,t} =3, S_{2,t} = 4,S_{3,t} =2,S_{4,t} =5, S_{5,t} =1$. This constraint can be explained that if the path from customer $i$ to $j$ is selected, i.e., $V_{i,j,t}(\omega)=1$, then $S_{i,t}(\omega)$ must be less than $S_{j,t}(\omega)$. 

In addition, the constraint in (\ref{e_order1}) uses $S_{i,t}(\omega)$ to ensure that all customers in the morning time window will be served before the afternoon time window. Similar to the constraint in (\ref{e_order1}), that in (\ref{e_order2}) ensures that all customers in the afternoon time window will be served before the evening time window.

The last three constraints in (\ref{e_bound1})-(\ref{e_bound3}) indicate the types and bounds of the decision variables. $X_{i,t}$, $W_{t}$, $Y_{i,r}(\omega)$, and $V_{u,v,t}(\omega)$ are binary variables, and $S_{i,t}(\omega)$ takes a value between one and the total number of customers.


\section{Performance Evaluation}
\label{s_performance}
\subsection{Parameter Setting}
We consider the system model with three types of FTL trucks and one LTL carrier. FTL trucks include a panel van, a 10ft box truck, and a 14ft box truck, the capacities of which are $F_1 = 1060$, $F_2 = 1360$ and $F_3 = 2268$ kilograms, respectively. Based on Singapore commercial vehicle rental agencies~\cite{ref_sgvehicle}, \cite{ref_vehicle}, the initial costs for renting a truck are set as $\bar{C}_{1} = S\$280, \bar{C}_{2} = S\$440$ and $\bar{C}_{3} = S\$640$. The cost of FTL truck routing is $\ddot{C}_{u,v} = K_{u,v} \times 1.05 \times 0.1$, which is calculated based on the distance between locations $u$ and $v$ multiplied by the approximate fuel price (S\$ per litre) and the average fuel consumption rate (litre per kilometer). We base the values of the parameters on that from Singapore Government statistics~\cite{ref_gov}. For the LTL carrier, we adopt the parameters from the Speedpost service offered by SingPost company~\cite{ref_singpost}. The LTL carrier service charge is $S\$21$ for a 30 kilogram package. $A_i$ is set equal to 30 kilograms. 

In this paper, we present the evaluation results from two different datasets including Solomon Benchmark Suite~\cite{ref_solomon} and Singapore road network. For Solomon Benchmark Suite, we adopt the file {\ttfamily C101} with some modification in our evaluation. The customer time window is calculated based on {\ttfamily READYTIME}$/150+9$ ($ 12<\mathcal{I}^{(m)} , 12\leq\mathcal{I}^{(a)}<15 , 15\leq\mathcal{I}^{(e)}$). For Singapore road network, we randomly choose 20 customer addresses and the location of the depot in Singapore. We assume that the traveling distance from location $u$ to location $v$ and from location $v$ to location $u$ are the same. The time windows are based on Solomon Benchmark Suite. 

For the presented experiments, we implement the stochastic integer programming model using GAMS Script~\cite{ref_gams}. Note that some parameters are varied for different experiment scenarios.

\subsection{Results and Explanations}

\begin{figure}
\center\includegraphics[width=0.45\textwidth]{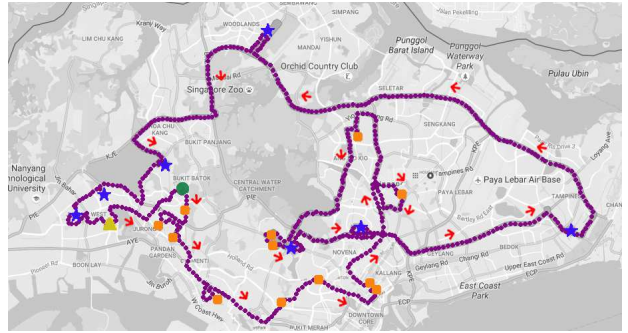} 
 \caption{The van vehicle route of the real Singapore Road Network dataset. 
 }
 \label{f_routeMap}
\end{figure}

\begin{figure*}

$\begin{array}{ccc} 
\hspace{-3.2em}
 \includegraphics[clip,trim=0cm 0cm 0cm 0,width=0.4\textwidth]{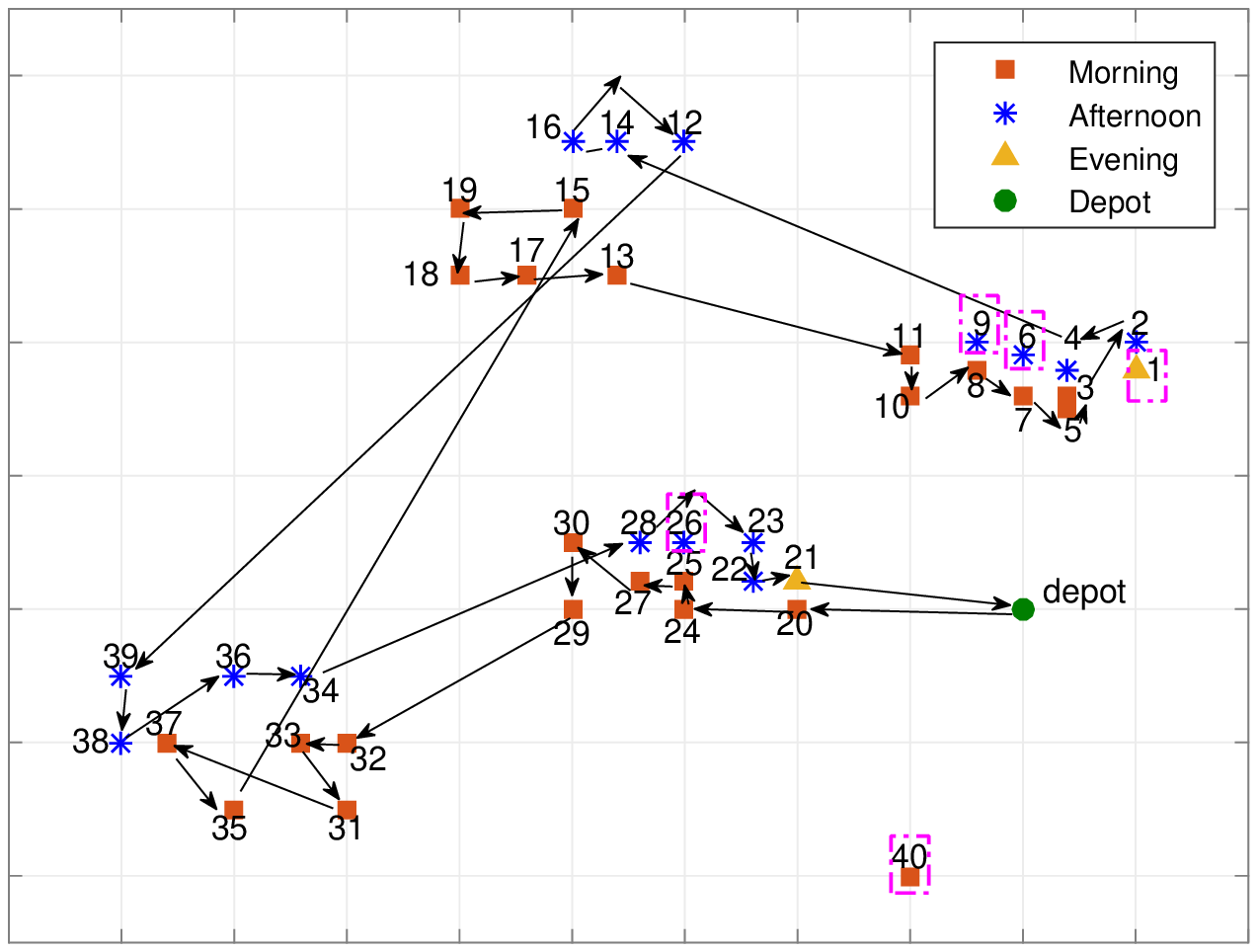} 
&\hspace{-2.8em}
\includegraphics[width=0.39\textwidth]{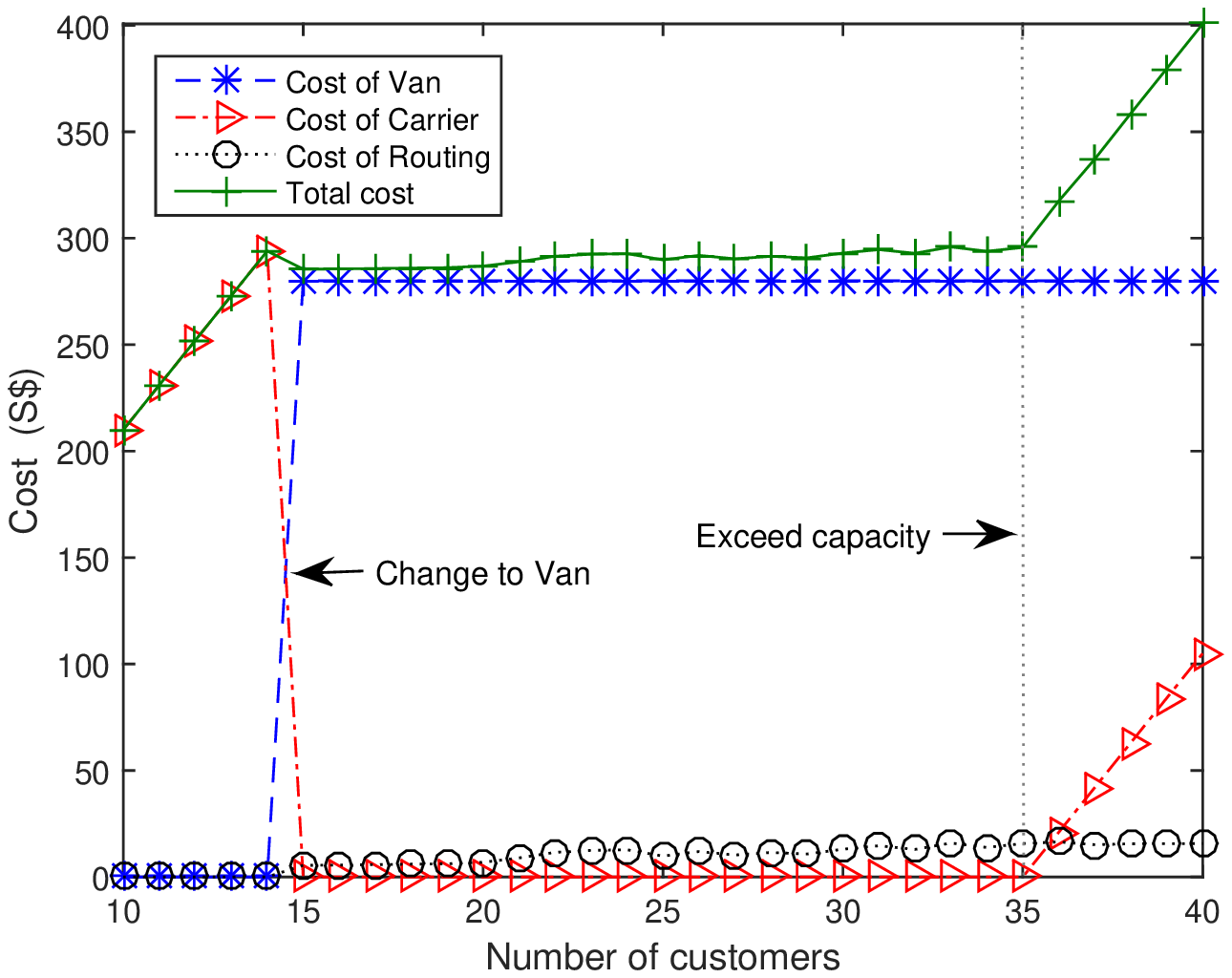} 
&\hspace{-2.8em}
\includegraphics[width=0.39\textwidth]{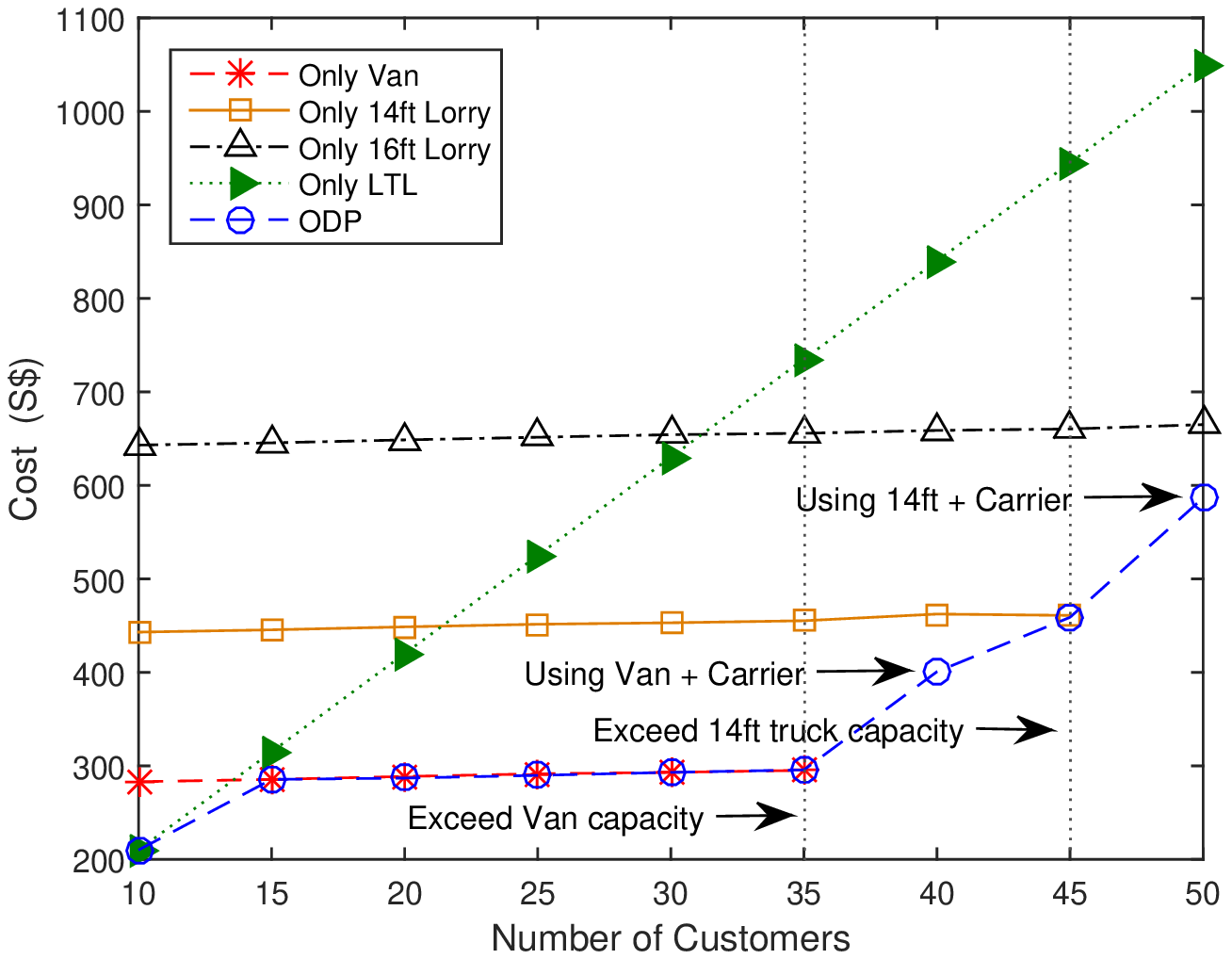}  \\
\end{array}$
 \minipage{0.3\textwidth}
 \caption{The van vehicle route of Solomon benchmark. 
}
 \label{f_routeSolo}
 \endminipage\hfill
\minipage{0.3\textwidth}
 \caption{Variant numbers of customers with one van truck in a system}
 \label{f_100p}
 \endminipage\hfill 
 \minipage{0.3\textwidth}
 \caption{The comparison  between using only carrier, only one kind of FTL trucks, and the ODP.
 }
 \label{f_modelcom}
 \endminipage
\vspace{-0.6cm}
\end{figure*}

The vehicle routing solutions are presented in Figure~\ref{f_routeMap} and Figure~\ref{f_routeSolo} for  Singapore road network (20 customers) and Solomon Benchmark Suite (40 customers), respectively. From the figures, only one panel van is selected. We observe that all customers in Singapore dataset are served by the FTL truck. However, for the Solomon Benchmark test result, 5 customers are served by the LTL carrier due to the capacity limit of the FTL truck. The total costs from the Solomon Benchmark and Singapore road map are S\$400.686 and S\$294.826. Note that the former includes S\$280 of truck initial cost, \$105 of LTL carrier service charge, and \$15.686 of routing cost. The latter includes only \$280 of truck initial cost and \$14.826 of routing cost.

\subsection{Impact of the Number of Customers}

We consider the case that all customers have demand, i.e., one scenario in this case. The total cost when the number of customers increases is presented in Figure~\ref{f_100p}. When the number of customers is few, e.g., less than 15, the supplier always serves all the customers by the LTL truck. This is because the initial cost of the FTL truck is more expensive. When the number of customers increases to more than 15, only the FTL truck is used since its initial cost and routing cost become cheaper than that using LTL carrier. The truck can handle up to 35 customers due to the capacity limit, i.e., a package of each customer is 30 kilograms.

\subsection{Comparison}

We next compare the different schemes including using only van, using only 14ft lorry, using only 16ft lorry, using only LTL carrier, and our proposed ODP. The results are shown in Figure~\ref{f_modelcom}. If the supplier uses only FTL trucks, each truck has an initial cost. When the number of customers increases, the total cost increases almost linearly due to routing cost. Moreover, each truck has a capacity limit, and the number of customers that can be served is confined by such a constraint. On the other hand, using the LTL carrier is more flexible, and the supplier only needs to pay according to the actual demand. However, the LTL carrier charge is expensive and the total cost increases sharply especially when the number of customers is large. Evidently, the proposed ODP based on stochastic integer programming achieves the lowest total cost. This is due to the fact the ODP always uses the cheapest option.

\subsection{Impact of Customer Locations}

\begin{figure}[h]
\hspace{-1.5em}
\vspace{-0.2cm}
\begin{center}
$\begin{array}{cc} 
\hspace{-1.4cm}
\includegraphics[width=0.3\textwidth]{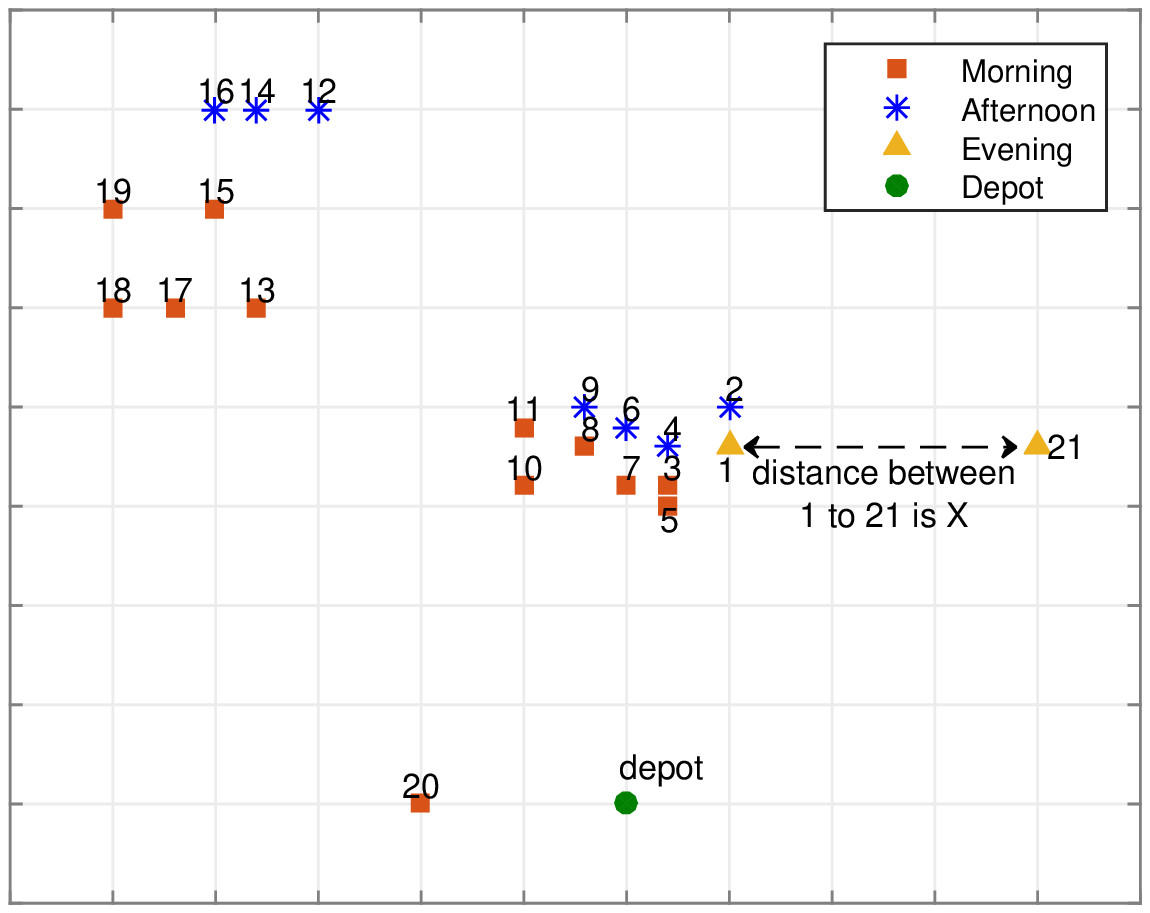}
&\hspace{-0.8cm}\includegraphics[width=0.3\textwidth]{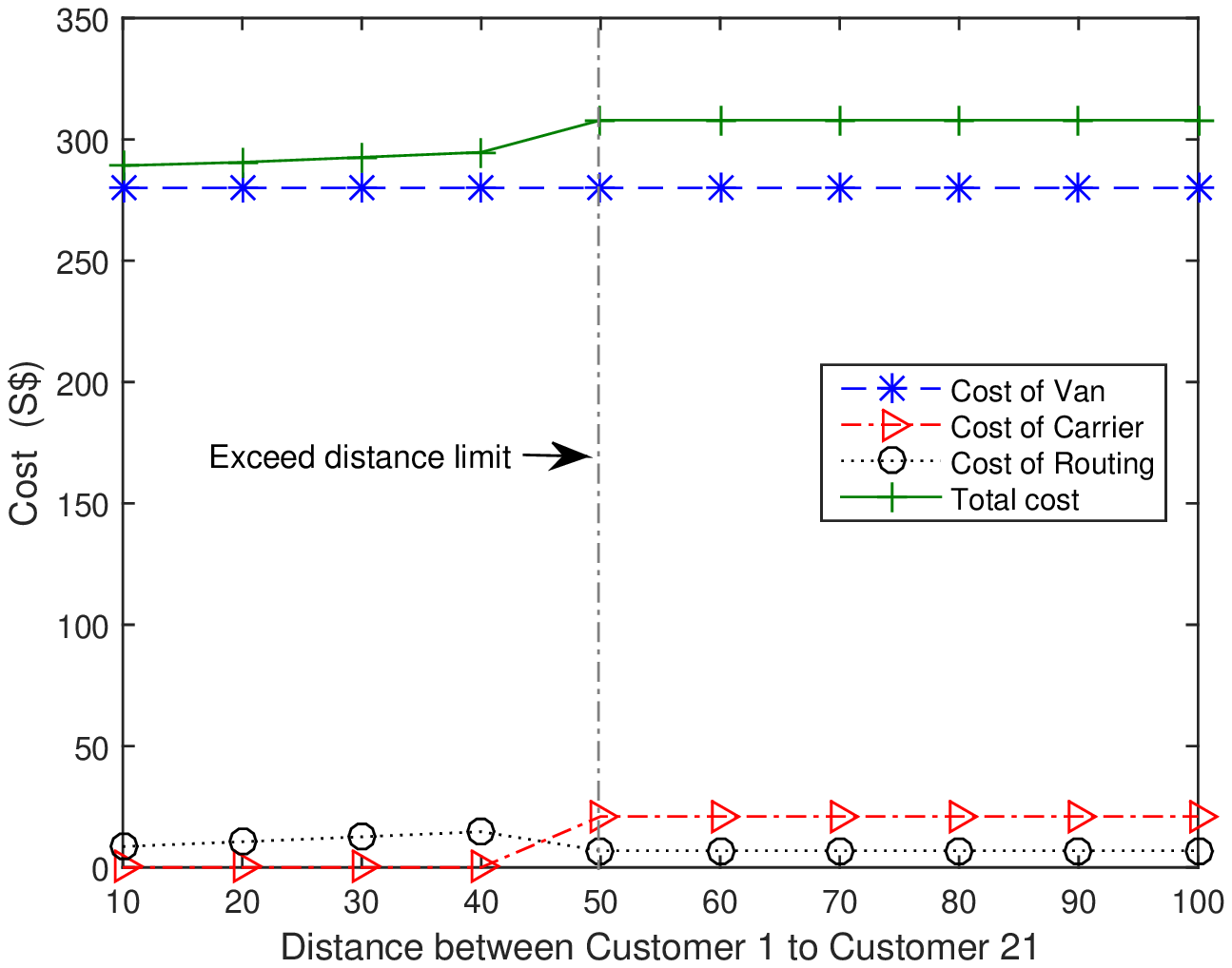}
\\\hspace{-1.4cm}(a)&
\hspace{-0.8cm} (b) 
\end{array}$
\end{center}
\vspace{-0.3cm}
\caption{Adding a customer ($C_{21}$) to various locations as presented in (a). (b) presents total delivery cost.}
\label{f_location}
\end{figure}

We then consider 21 customers, where customers $C_1$ to $C_{20}$ are from Solomon Benchmark dataset. The location of customer $C_{21}$ is varied. The locations of all customers in this experiment are shown in Figure~\ref{f_location}(a). Figure~\ref{f_location}(b) presents the total cost. When customer $C_{21}$'s location becomes farther away from $C_1$, the truck routing cost to this customer increases. Until the distance is more than the traveling distance limit, i.e., 50 kilometers, the supplier uses the LTL carrier to serve this customer $C_{21}$. Otherwise, the customer may not be served during the time window. 
\vspace{-0.8cm}

\section{Conclusion}
In this paper, we have proposed the Optimal Delivery Planning (ODP) for a supplier to make the best decisions of full-truckload and less-than truckload delivery. However, the delivery demand from customers is random and the information of which is not known when the trucks have to be reserved. Therefore, we have formulated the ODP as the two-stage stochastic programming with customer demand uncertainty. The ODP optimizes the total delivery cost for the supplier. The trade-off between truck allocation and carrier assignments has been optimized. The experiment results from two datasets, i.e., Solomon Benchmark and Singapore road network, have been presented. Compared to the other baseline schemes, the ODP has successfully achieved the lowest delivery cost. For the future work, we will incorporate a truck break down event into the optimization model. 

\section{Acknowledgment}
This work is partially supported by Singapore Institute of
Manufacturing Technology-Nanyang Technological University
(SIMTech-NTU) Joint Laboratory and Collaborative research
Programme on Complex Systems.


\end{document}